\documentclass{scrartcl}
\usepackage[utf8]{inputenc}
\usepackage[T1]{fontenc}
\usepackage{graphicx}
\usepackage{grffile}
\usepackage{longtable}
\usepackage{wrapfig}
\usepackage{rotating}
\usepackage[normalem]{ulem}
\usepackage{amsmath}
\usepackage{textcomp}
\usepackage{amssymb}
\usepackage{capt-of}
\usepackage{hyperref}
\usepackage{xcolor}
\usepackage{tikz}
\usepackage{soul}
\usepackage{listings}
\usepackage[version=3]{mhchem}
\usepackage{doi}
\usepackage{amsmath}
\usepackage[british, english]{babel}
\hypersetup{colorlinks=true,citecolor=red!80!black,urlcolor=blue!60!black}
\usepackage{algorithm}
\usepackage{algpseudocode}
\let\sc\scshape
\lstdefinelanguage{Julia}%
{morekeywords={abstract,Array,break,case,catch,const,continue,do,else,elseif,%
end,export,false,Float64,for,function,immutable,import,importall,if,in,%
macro,module,mutable,new,otherwise,quote,print,return,struct,switch,true,try,type,typealias,%
using,while},%
sensitive=true,%
alsoother={$},%
morecomment=[l]\#,%
morecomment=[n]{\#=}{=\#},%
morestring=[s]{"}{"},%
morestring=[m]{'}{'},%
stringstyle=\color{gray},%
}[keywords,comments,strings]%

\usepackage[backgroundcolor=yellow!10!white]{mdframed}
\tikzstyle{unit}=[rectangle, draw=blue!80!black, fill=black!5!white]
\tikzstyle{stream}=[rectangle, draw=white, fill=yellow!50!white]
\usepackage{newunicodechar} 
\newunicodechar{α}{\ensuremath{\alpha}}
\newunicodechar{δ}{\ensuremath{\delta}}
\newunicodechar{Δ}{\ensuremath{\Delta}}
\newunicodechar{∈}{\ensuremath{\in}}
\newunicodechar{±}{\ensuremath{\pm}}
\definecolor{verypaleblue}{rgb}{0.95,0.95,1}

\lstdefinelanguage{Maxima}%
{morekeywords={define,diff,print,solve},%
sensitive=true,%
alsoother={$},%
morecomment=[l]\#,%
morecomment=[n]{\#=}{=\#},%
morestring=[s]{"}{"},%
morestring=[m]{'}{'},%
}[keywords,comments,strings]%

\usepackage[overlay]{textpos} \TPGrid[0pt,0pt]{20}{20}
\author{Eric S. Fraga\\ \href{mailto:e.fraga@ucl.ac.uk}{\texttt{e.fraga@ucl.ac.uk}} \\ Sargent Centre for Process Systems Engineering\\ Department of Chemical Engineering\\ University College London (UCL)}
\date{\today}
\title{Multiple simultaneous solution representations in a population based evolutionary algorithm}
\hypersetup{
 pdfauthor={Eric S Fraga\\ \href{mailto:e.fraga@ucl.ac.uk}{\texttt{e.fraga@ucl.ac.uk}} \\ Sargent Centre for Process Systems Engineering\\ Department of Chemical Engineering\\ University College London (UCL)},
 pdftitle={Multiple simultaneous solution representations in a population based evolutionary algorithm},
 pdfkeywords={},
 pdfsubject={},
 pdfcreator={Emacs 28.0.50 (Org mode 9.4.6)}, 
 pdflang={English}}
\begin{document}

\maketitle
\begin{abstract}
The representation used for solutions in optimization can have a significant impact on the performance of the optimization method.  Traditional population based evolutionary methods have homogeneous populations where all solutions use the same representation.  If different representations are to be considered, different runs are required to investigate the relative performance.  In this paper, we illustrate the use of a population based evolutionary method, Fresa, inspired by the propagation of Strawberry plants, which allows for multiple representations to co-exist in the population.

Fresa is implemented in the Julia language.  Julia provides dynamic typing and multiple dispatch.  In multiple dispatch, the function invoked is determined, dynamically at run time, by the types of the arguments passed to it.  This enables a generic implementation of key steps in the plant propagation algorithm which allows for a heterogeneous population.  The search procedure then leads to a competition between representations automatically.

A simple case study from the design of operating conditions for a batch reactor system is used to illustrate heterogeneous population based search.
\end{abstract}
\section{Introduction}
\label{sec:org108d63e}
The representation of possible solutions to a problem is a key element in the application of evolutionary algorithms for optimization problems \cite{10.1145/1570256.1570415}.  The choice of representation defines the search domain and the operators that create new points in the search are based on the representation.  Alternative representations may lead to different outcomes in the search \cite{fraga-etal-2018a} although algorithms may be adapted to perform equally well for different representations in some cases \cite{687882}.  Evolutionary algorithms that use operators that combine two or more points from the population to create new points, e.g. crossover in a genetic algorithm, will typically require that the population be homogeneous, i.e. consist of points all using the same representation.  This restriction motivates, for instance, the use of metameric representations \cite{ryerkerk-etal-2019}.  In some cases, metameric representations may have implicit biases which will affect the efficiency or effectiveness of the search methods \cite{10.1007/BFb0040779}.

For evolutionary methods which do not use operators which combine multiple points for the creation of new points, heterogeneous populations including the simultaneous presence of alternative representations may be considered.  This paper describes an evolutionary method, based on plant propagation, which enables the use of heterogeneous populations.  A simple case study on the design of operating temperature profiles for a batch reactor system is presented.  Two alternative representations for the temperature profiles are considered.

Although multiple representations could be evaluated by solving the problem multiple times with homogeneous populations, one of the attractions of heterogeneous populations is the ease with which the multiple representations can be compared.  In a single run, the solutions with different representations compete against each other.  To highlight the ease with which this can be done, the code for the case study presented below is included in Appendix \ref{code}.
\subsection{Nomenclature}
\label{sec:org2c2df14}
In the following, we use this nomenclature to describe the search for an optimum:
\begin{description}
\item[{point}] A solution to the problem where different points may use different representations.
\item[{domain}] Represented by \(\cal{D}\) and which is dependent on the representation used.
\item[{problem}] An optimization problem which can generally be described as
\begin{equation}
\label{eq-problem}
  \min_{x \in {\cal D}(x)} z = f(x)
\end{equation}
and
\[g(x) \le 0 \]
where \({\cal D}(x)\) which, through an abuse of notation, indicates that the search domain depends on the representation used for \(x\), \(f(x)\) is the objective function, and \(g(x)\) are constraints including, without loss of generality, both inequality and equality constraints.
\end{description}
\section{A plant propagation meta-heuristic algorithm}
\label{sec:orgae8bc2d}

\begin{figure}[hbtp]
\renewcommand{\algorithmicrequire}{\textbf{Given: }}
\renewcommand{\algorithmicensure}{\textbf{Returns: }}
\begin{algorithmic}[1]
  \Require $f(x)$, \(p_0\), ${\cal D}$, $n_g$, $n_p$, $n_r$.
  \Ensure $z$, the best solution found for single objective problems or the set of non-dominated points for multi-objective problems.
  \State $p \gets $ initial population \(p_0\)
  \For {$n_g$ generations}
  \State $N \gets $ fitness($p$)
  \Comment evaluate and rank
  \If {elitism}
  \State $\tilde{p} \gets \left\{ \text{most fit individual(s) in }p\right \} $
  \Comment Single solution or non-dominated set
  \Else
  \State $\tilde{p} \gets \emptyset$
  \EndIf
  \For {$i \gets 1, \ldots, \min\left\{n_p,\left|p\right|\right\}$}
  \Comment propagate up to $n_p$ individuals
  \State $j \gets $ select($p,N$)
  \Comment fitness based selection
  \State $\tilde{p} \gets \tilde{p} \, \cup \left \{ x_j \right \} \, \cup \left\{x_k | x_k = \text{neighbour}(x_j,N_j); k=1,\ldots,n_r\propto N_j\right\}$
  \Comment runners
  \EndFor
  \State $p \gets \tilde{p}$
  \Comment population for new generation
  \EndFor
  \State $z \gets $ most fit individual(s) in final population.
\end{algorithmic}
\caption{\label{fresa-algorithm}The Fresa plant propagation algorithm}
\end{figure}

Fresa \cite{fresa} is a \emph{variable neighbourhood search} meta-heuristic method inspired by the propagation of Strawberry plants using runners \cite{salhi-fraga-2011a}.  Fresa has recently been used for solving problems in dynamic optimization that arise in the process industries \cite{rodman-etal-2018a,fraga-2019a}.  The algorithm is presented in Figure \ref{fresa-algorithm}.  For this algorithm, the inputs and parameters required are described in Table \ref{tab-parameters}.  The attraction of Fresa is its simplicity, leading to efficient implementation, including the use of multi-threading for the evaluation of new points in parallel when the objective function is computationally expensive.  Also, the algorithm has few tunable parameters and, as shown in a recent study \cite{dejonge-vandenberg-2020a}, the performance of the method is not highly sensitive to the values of these parameters.  

\begin{table}[hbtp]
\caption{\label{tab-parameters}Summary of specifications defining the problem to solve and the parameters required by Fresa.}
\centering
\begin{tabular}{lp{8cm}}
Specification & Description\\
\hline
\(f(x)\) & The objective function(s) to be evaluated at a point \(x\) in the search domain.\\
 & \\
\(p_0\) & Initial population of points.\\
 & \\
\({\cal D}\) & The search \emph{domain}, implemented as a pair of functions, \emph{lower} and \emph{upper} which will be evaluated with actual points in the population.\\
 & \\
\(n_g\) & Number of generations to perform during the search.\\
 & \\
\(n_p\) & Maximum number of points in the population to choose for propagation for each generation.\\
 & \\
\(n_r\) & Maximum number of runners to create for any point propagated. Defaults to 5.\\
\end{tabular}
\end{table}

\subsection{Generic search}
\label{sec:org39c1b36}

The basic evolutionary procedure in this plant propagation algorithm is that the most fit points generate more but shorter runners and less fit points generate fewer but longer runners.  This provides a balance between the \emph{exploitation} of good points and the \emph{exploration} of the search space.  The implementation of the propagation, the creation of runners, is delegated to the \texttt{neighbour} function, line 11 in Algorithm \ref{fresa-algorithm}.  This function takes two arguments: the point to propagate and its fitness.  The propagation of a single point does not depend on any other points explicitly and only implicitly in that the fitness is relative to the full population.

Fresa makes no assumptions of the representation used to search the space of possible solutions or the search space itself.  The former are implicit in the members of the population and the latter is implicitly defined by the \texttt{neighbour} function.  Fresa achieves this generic or agnostic behaviour through implementation in the Julia language\footnote{\url{https://julialang.org/}} \cite{bezanson-etal-2017a}, although almost any programming language could be used in principle.  Using Julia facilitates the agnostic behaviour through its support for dynamic typing and multiple dispatch.  These features enable the development of generic software methods that can subsequently be applied to problems not originally envisaged by the authors of the software.

In the case of Fresa, the generic aspects of the implementation are in the definitions of the decision variables which represent the points in the search domain and the results of evaluating those decision variables, the values returned by the objective function.  As the neighbour function can differ depending on the point to be propagated, multiple dispatch with dynamic typing enables Fresa to have a heterogeneous population where different representations of points in the search domain (and the search domain itself, for that matter) can be considered simultaneously.

\section{Case study: batch reactor operating temperature profile}
\label{sec:org789d83f}
We consider the problem of identifying the best operating temperature profile for the control of a batch reactor system.  In this problem, the underlying system is a batch reactor with the following reactions taking place \cite{cizniar-etal-2006a}:
\begin{equation*}
  \ce{A -> B -> C}
\end{equation*}
where the aim is to produce species B from species A.  Unfortunately, there is a second reaction which consumes B to produce an undesired species C.  The rates at which the individual reactions take place are affected by the \emph{operating temperature} and choosing this temperature, as a function of time, is a key operating design decision.

The system of reactions is modelled as a system of ordinary differential equations with the amounts of the species present at any time described by their concentrations as the dependent variables in these equations:
\begin{align*}
  \frac{\text{d}}{\text{d}t} C_A &= -k_1 C_A^2 \\
  \frac{\text{d}}{\text{d}t} C_B &= k_1 C_A^2 - k_2 C_B \\
  \frac{\text{d}}{\text{d}t} C_C &= k_2 C_B
\end{align*}
for \(t \in [0,1]\).  The temperature dependent rates of the two individual reactions are given by
\begin{align*}
  k_1 &= 4000 e^{-\frac{2500}{T}} \\
  k_2 &= 620\,000 e^{-\frac{5000}{T}}
\end{align*}
\(T(t)\) is the time varying operating temperature.  As a result, \(k_1\) and \(k_2\) also vary with time. The operating temperature profile is to be determined so as to achieve the best outcome.  This defines an optimization problem:
\begin{equation}
\label{optimization-problem}
\max_{T(t)} z = C_B(1)
\end{equation}
where \(C_B(1)\) is the concentration of B at the final time \(t=1\).
\subsection{Representing the temperature profile}
\label{sec:org85fdb56}
In addressing a similar, albeit more complex, design problem \cite{rodman-etal-2018a}, it was noted that there exist alternative representations for the operating temperature profile.  For instance, a profile could be defined by a set of discrete points in the time domain, with a temperature value to be determined at each time point.  The resulting profile would then be piece-wise linear.  Alternatively, the profile could be described by a small number of spline curves, defined by their end-points and starting and finishing derivatives.  The advantages of the latter representation is that the space of designs, the domain for the search, consists of profiles that are more realistic in terms of operation and control and these profiles have more \emph{pleasing} shapes.  Both of these features may aid the adoption of proposed operating process designs by the operating engineers.

In detail, the piece-wise linear representation consists of an initial temperature, \(T_0\), followed by a set of time points and the change in temperature to achieve in the interval defined by each time point.  Each time point is a value in \([0,1]\) which specifies the fraction of the remaining time for the size of this time interval.  The temperature change is a value in \([-1,1]\) which, when multiplied by \(\Delta T_{\max}\), specifies the change in temperature over this time interval.  Therefore, if \(n_t\) time intervals are desired, the encoding is \(2\times n_t + 1\) real values.  All possible encoded values correspond to feasible solutions.  For the case study, \(n_t = 4\) so the encoding has 9 real values.

The quadratic spline representation is based on two quadratic polynomials.  The initial and final temperatures are specified as is the time point at which the two polynomials meet.  With the conditions that the slope of the polynomials be 0 at the initial time and at the final time and that the slopes of the two polynomials be the same at the meeting point, the three values are sufficient to define the temperature profile.  This representation has been used previously in a study for the solution to dynamic (i.e. time varying) multi-objective optimization problems \cite{fraga-2019a}.  

\subsection{Evolving a heterogeneous population}
\label{sec:org56154ac}
In Fresa, the domain for the search procedure depends on the representation as does the identification of neighbouring points for propagation.  By providing functions with the appropriate \emph{signature}, or by interrogating the point for its \emph{type}, the appropriate domain or neighbour function can be identified or automatically invoked.  If the initial population given to Fresa contains points with different representations, these will propagate and lead to populations with a varying mix of representations unless, of course, selection pressures lead to one or more representations dying out by not being selected for propagation.

We, therefore, consider solving the problem in equation \ref{optimization-problem} with an initial population consisting of points with both representations: a piece-wise linear representation for the operating temperature profile and a two segment quadratic spline profile.  The parameters for Fresa are 100 generations, \(n_g=100\), select 5 points each generation for propagation, \(n_p=5\), and create a maximum of 5 runners for each point propagated, \(n_r=5\).  No elitism is used so a new generation will consist of those points selected for propagation together with the points resulting from the propagation (cf. lines 7 and 11 of Algorithm \ref{fresa-algorithm}).

\begin{figure}[hbtp]
\centering
\includegraphics[width=0.9\linewidth]{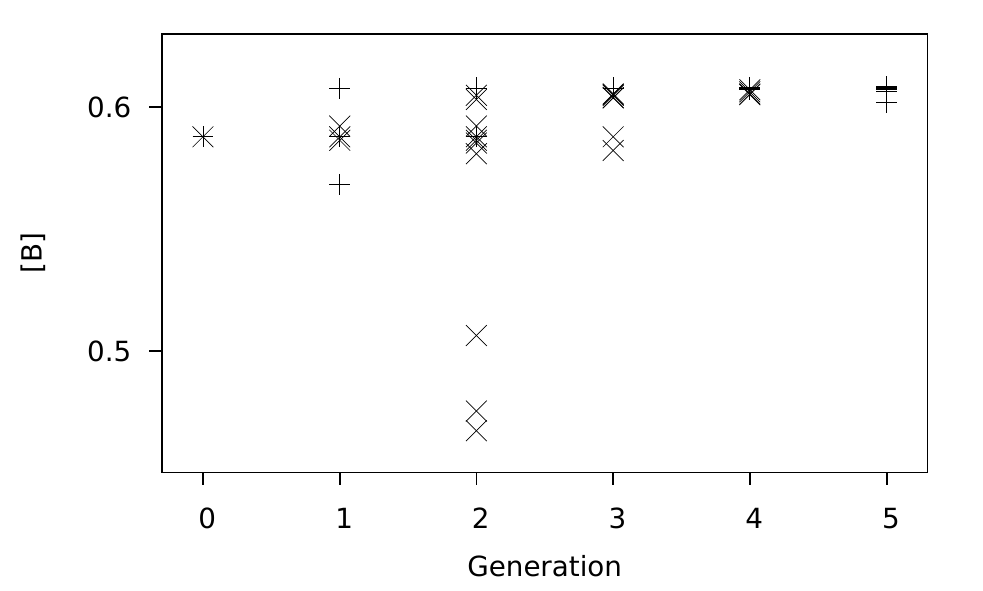}
\caption{\label{evolution-plot-figure}Evolution of population with 2 representations: \(+\) for quadratic spline and \(\times\) for piece-wise linear.  The initial population consists of 2 members, one with each representation, both of which result in the same objective function value.}
\end{figure}

Figure \ref{evolution-plot-figure} shows the make-up of the population after each of the first 5 generations in one of the attempts at solving the problem, starting with an initial population comprising two points, one for each representation, with both points having the same objective function value.  In this figure, we see that points with both representations co-exist in the population until generation 4.  In the fifth generation, only those using the quadratic spline representation remain.  At this point, the best solutions have already achieved close to optimum solutions with a concentration of just above 0.6 for species B.

\begin{table}[hbtp]
\caption{\label{stochastic-outcome-table}Summary of the behaviour of the population with competing representations, showing final outcome and when homogeneity was reached in the population for 10 attempts at the problem, sorted by final outcome in terms of objective function value.}
\centering
\begin{tabular}{lrr}
Final representation & Homogeneous generation & Best found\\
\hline
Quadratic spline & 7 & 0.6091\\
Quadratic spline & 5 & 0.6088\\
Quadratic spline & 6 & 0.6087\\
Piece-wise linear & 7 & 0.6085\\
Piece-wise linear & 10 & 0.6083\\
Piece-wise linear & 4 & 0.6080\\
Piece-wise linear & 8 & 0.6080\\
Piece-wise linear & 13 & 0.6074\\
Piece-wise linear & 8 & 0.6069\\
Piece-wise linear & 11 & 0.6069\\
\end{tabular}
\end{table}

The stochastic nature of Fresa means that different attempts at solving the problem can potentially lead to not only quantitatively different outcomes, e.g. the best objective function value found, but also qualitatively different outcomes, for instance in terms of the representations present in the final population.  Table \ref{stochastic-outcome-table} summarises the results of attempting the problem 10 times with the same initial population.  The entries have been ordered by the value of the objective function, the third column, for the best solution found in each run.  The first column indicates which representation dominated and the second column shows at which generation the population became homogeneous in terms of representation.  In most cases, when the quadratic spline representation loses out to the piece-wise linear representation, this happens later than when the converse happens.

What the table further shows is that the best outcomes happen when the quadratic spline representation \emph{wins out} over the piece-wise linear representation.  However, this happens only 3 times out of the 10 runs.  One possible reason for the quadratic spline representation doing better in terms of objective function value is that the representation is simpler with three values defining the profile: start temperature, time point for where the two quadratics meet, and the final temperature.  The piece-wise linear profile requires 9 values: initial temperature and four linear segments (duration and change in temperature incurred).  However, it is unclear why the piece-wise linear representation is more successful in dominating the population.  Further research into selection pressure will be useful here.
\subsection{A multi-objective version of the case study}
\label{sec:orgaa8b774}

In the previous section, the single objective problem, equation \ref{optimization-problem}, was considered.  In this section, we extend the problem to one with two criteria:

\begin{equation}
\label{multi-objective-optimization-problem}
\min_{T(t)} z = \left [ \begin{array}{r} - C_B(1) \\ C_C(1) \end{array} \right ]
\end{equation}
where \(C_B(1)\) is the concentration of B at the final time and \(C_C(1)\) the concentration of C at the final time \cite{fraga-2019a}.  The first objective is the negated value of the concentration of B to make both objectives have the same target: minimization.  This multi-objective problem retains the desire to maximise the concentration of B but at the same time minimizing the final concentration of C.  This would be a reasonable goal when the effluent of the reactor may be sent back into the next batch to consume any unreacted amounts of species A.

In a multi-objective problem, the definition of fitness of points in a population can be defined in a number of ways.  One popular approach is to assign points in a population a fitness based on the recursive identification of points which are non-dominated \cite{deb-2000}.  Alternatively, points can be ranked by combining the rankings with respect to the individual criteria, for example using a Hadamard product of the individual rankings \cite{fraga-amusat-2016a}.  The latter emphasises points that are towards the ends of the approximation to the Pareto frontier.  A variation on this latter approach is to use a Borda sum of the individual rankings.

All three fitness approaches have been considered.  For the non-dominated and the Borda sum rankings, points using the piece-wise linear representation disappear from the population in a few generations every time.  For the Hadamard product ranking, points with both representations typically survive until the final generation when considering 100 generations with 20 points propagated at each generation.

\begin{figure}[hbtp]
\centering
\includegraphics[width=0.9\linewidth]{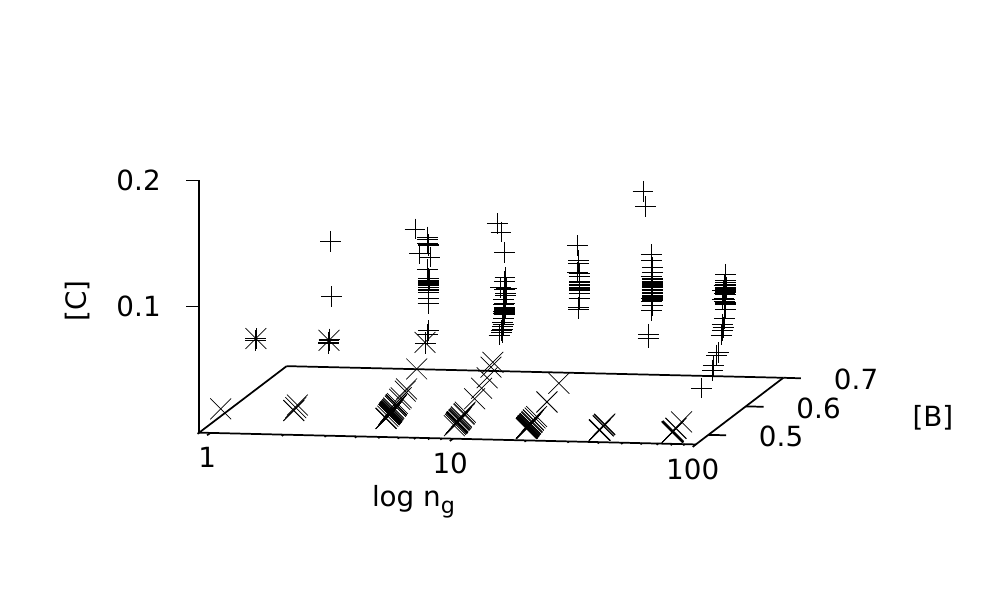}
\caption{\label{multi-objective-evolution-figure}Evolution of population for the multi-objective case with + indicating points represented by quadratic splines and \texttimes{} for piece-wise linear representation.}
\end{figure}

Figure \ref{evolution-plot-figure} shows the evolution of the population starting with the same initial guesses as before: one point for each alternative representation with each point resulting in the same objective function values.  Recall that the aim is to maximise the production of species B while minimising the amount of C present.  Therefore, the utopia point is \([1.0,0.0]\) for the concentrations of B and C respectively.  The set of non-dominated points will be a curve starting at the bottom at the front of the plot in the figure (i.e. towards the generation axis) and curving up towards the back.  As the number of generations, \(n_g\) increases, the breadth of the approximation to the Pareto frontier covered by points using the quadratic spline representation increases and the breadth of the frontier covered by points with the piece-wise linear representation decreases.  

\begin{figure}[hbtp]
\centering
\includegraphics[width=0.9\linewidth]{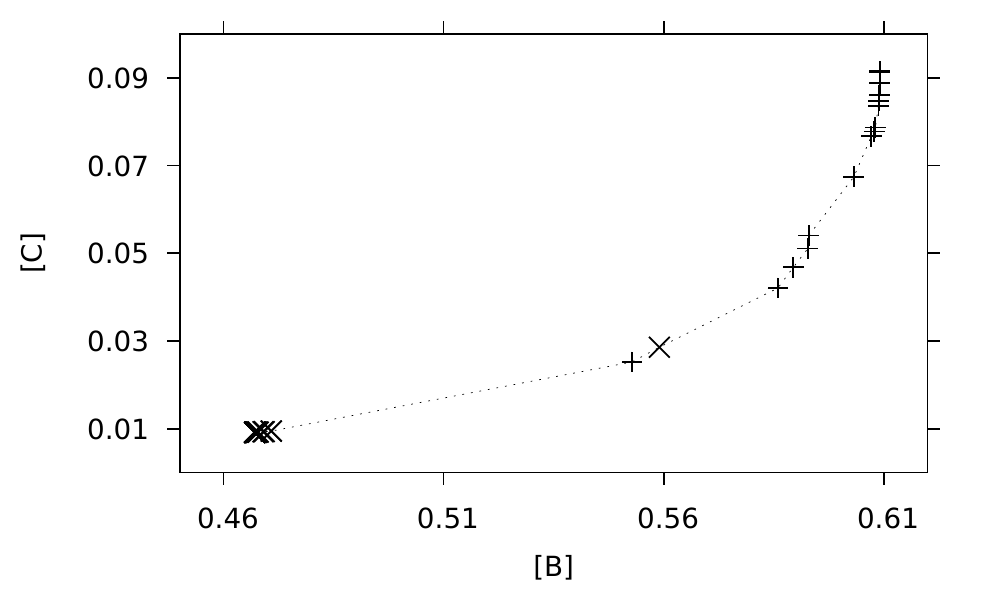}
\caption{\label{multi-objective-final-pareto-set-figure}Final set of non-dominated points for the multi-objective case study.  Both representations are present in the final set of non-dominated points with piece-wise linear representations denoted by \texttimes{} and quadratic spline representations by \(+\).  The fitness function for selection was based on the Hadamard product of the rankings of points with respect to the individual criteria \cite{fraga-amusat-2016a}.}
\end{figure}

If further generations were to be performed, it would seem likely to conclude that only points with quadratic splines may be present. However, Figure \ref{multi-objective-final-pareto-set-figure} presents the set of non-dominated points in the final population.  Note that one point using the piece-wise linear representation has appeared in the middle of the approximation to the Pareto frontier.  Other than this point, however, there is a clustering of points using the piece-wise linear representation towards the left end point of the approximation to the Pareto frontier. The quadratic spline points are more dispersed and cover a greater part of the frontier approximation.  

\begin{figure}[hbtp]
\centering
\includegraphics[width=0.9\linewidth]{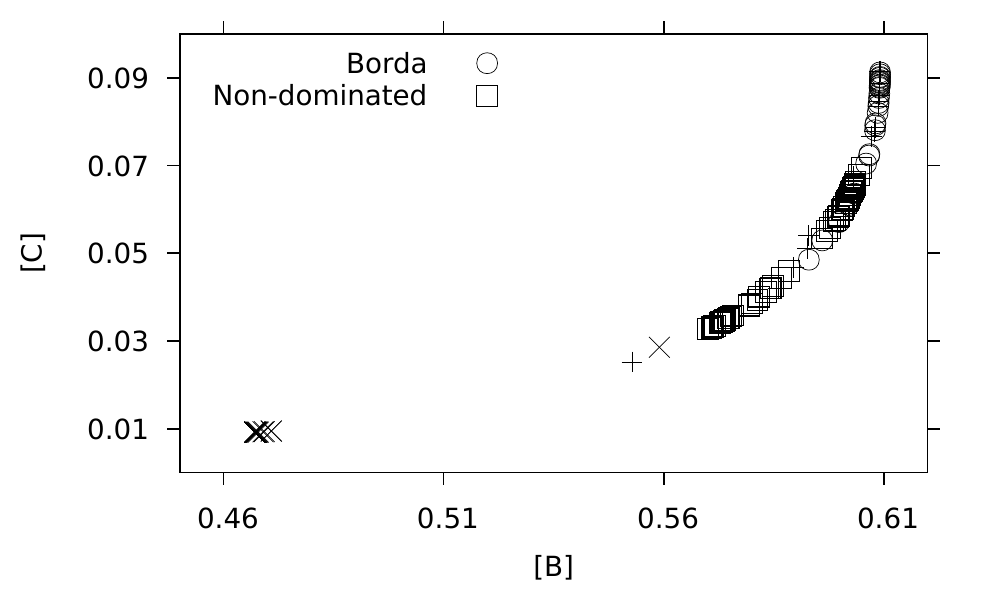}
\caption{\label{multi-objective-final-pareto-all-figure}Comparison of sets of non-dominated points in the final population for the three different fitness ranking approaches.  The points found using the Borda and non-dominated fitness ranking methods all correspond to quadratic spline representations and these have been superimposed on the plot showing the results using a Hadamard fitness ranking, Figure \ref{multi-objective-final-pareto-set-figure}.}
\end{figure}

The clumping is due to the Hadamard product fitness function \cite{fraga-amusat-2016a}.  Figure \ref{multi-objective-final-pareto-all-figure} superimposes the sets of non-dominated points obtained using the non-dominated ranking method \cite{deb-2000} and the Borda sum approach.  The non-dominated ranking method tends to find points in the middle of the frontier; the Borda method ends up with points all towards the right end point of the Pareto frontier approximation.  The Hadamard product ranking appears to perform best for this problem in the sense of covering more of the approximation to the Pareto frontier.

\section{Conclusions}
\label{sec:org99963d2}
A plant propagation algorithm (PPA) can evolve a population without the use of operators which require information about the representation of other points in the population.  Therefore, points with different representations may co-exist in the population and evolve independently.  By initialising the population with points using alternative representations, the evolutionary algorithm may propagate points with different representations with different probabilities.  Competition, in the sense of selection pressure, will then decide whether the representations may continue to be used as the population evolves.

This paper has presented results using Fresa, an implementation of a PPA in the Julia language.  Using the multiple dispatch and dynamic typing features of the Julia programming language facilitates the use of multiple representations for solutions in an evolutionary optimization procedure.  The case study, the design of the operating temperature profile for a batch reactor, is one which lends itself to different representations for alternative designs.  Two representations are considered: a piece-wise linear representation and one using a pair of quadratic spline polynomials.  Both single objective and multi-objective versions of the design problem have been solved.

For the single objective problem, although the piece-wise linear representation wins out more often, the best solutions are obtained with quadratic splines.  This motivates further research into selection and possibly the use of elitism in single objective cases, e.g. considering multiple elite sets, one for each representation.  For the multi-objective formulation, the results favour points using the quadratic spline representation although the behaviour depends on the fitness ranking method used.  This further motivates investigating fitness ranking methods with the aim of improving the coverage of the approximation to the Pareto frontier.

\bibliographystyle{acm}

\appendix
\section{The code \label{code}}
\label{sec:orgb586117}
The paper has been written with literate programming, using \texttt{org} mode\footnote{\url{https://orgmode.org/}} in the Emacs\footnote{\url{https://www.gnu.org/software/emacs/}} editor.  The results presented are from the code included directly in the paper by \emph{tangling} and running this code \cite{schulte-etal-2011a}.  All supporting code, including the Fresa\footnote{\url{https://github.com/ericsfraga/Fresa.jl}} \cite{fresa} plant propagation algorithm implementation and the batch reactor case study\footnote{\url{https://gitlab.com/ericsfraga/BatchReactor.jl}} are freely available as open source.

\subsection{Single objective case study}
\label{sec:org48178a7}
\lstset{language=julia,label=case-study-single,caption= ,captionpos=b,numbers=none}
\begin{lstlisting}
using BatchReactor
using Fresa
# define the objective function which is the concentration of the
# second species; as Fresa solves minimization problems and our
# problem is one of maximizing the concentration, we return the
# negated value.
function objective(profile :: BatchReactor.TemperatureProfile)
    results = BatchReactor.simulation(profile)
    ([-results[2]], 0.0)
end
# define the domain for the search procedure.  The domain is defined
# by lower and upper bounds but these depend on the actual type of
# profile considered
function lower(p :: BatchReactor.TemperatureProfile)
    if typeof(p) == BatchReactor.PiecewiseLinearProfile
        BatchReactor.PiecewiseLinearProfile(zeros(length(p.ft)),
                                            -ones(length(p.fT)),
                                            BatchReactor.Tmin)
    else
        BatchReactor.QuadraticSplineProfile(BatchReactor.Tmin,
                                            BatchReactor.Tmin,
                                            0.25)
    end
end
function upper(p :: BatchReactor.TemperatureProfile)
    if typeof(p) == BatchReactor.PiecewiseLinearProfile
        BatchReactor.PiecewiseLinearProfile(ones(length(p.ft)),
                                            ones(length(p.fT)),
                                            BatchReactor.Tmax)
    else
        BatchReactor.QuadraticSplineProfile(BatchReactor.Tmax,
                                            BatchReactor.Tmax,
                                            0.75)
    end
end
domain = Fresa.Domain(lower, upper)
# create initial population consisting of one initial profile for each
# type of representation
p0 = [Fresa.createpoint(BatchReactor.QuadraticSplineProfile(323.0,
                                                            323.0,
                                                            0.5),
                        objective)
      Fresa.createpoint(BatchReactor.PiecewiseLinearProfile([0.5,
                                                             0.5,
                                                             0.5,
                                                             0.5],
                                                            [0.0,
                                                             0.0,
                                                             0.0,
                                                             0.0],
                                                            323.0),
                        objective )]
# invoke Fresa
best, population = Fresa.solve(
    # the first 3 arguments are required
    objective,              # the objective function
    p0,                     # an initial point in the design space
    domain;                 # search domain for the decision variables
    # the rest are option arguments for Fresa
    elite = false,          # elitism by default
    ngen = 100,             # number of generations
    npop = 5,               # population size
    nrmax = 5,              # number of runners maximum
    populationoutput = true # output full population every generation
    )

println("Best solution: $best")
println("Full population:")
println("$population")
\end{lstlisting}

\subsection{Multi-objective case study}
\label{sec:orgf43ca50}
\lstset{language=julia,label=case-study-multiple,caption= ,captionpos=b,numbers=none}
\begin{lstlisting}
using BatchReactor
using Fresa
# define the objective function which is the concentration of the
# second species; as Fresa solves minimization problems and our
# problem is one of maximizing the concentration, we return the
# negated value.
function objective(profile :: BatchReactor.TemperatureProfile)
    results = BatchReactor.simulation(profile)
    ([-results[2], results[3]], 0.0)
end
# define the domain for the search procedure.  The domain is defined
# by lower and upper bounds but these depend on the actual type of
# profile considered
function lower(p :: BatchReactor.TemperatureProfile)
    if typeof(p) == BatchReactor.PiecewiseLinearProfile
        BatchReactor.PiecewiseLinearProfile(zeros(length(p.ft)),
                                            -ones(length(p.fT)),
                                            BatchReactor.Tmin)
    else
        BatchReactor.QuadraticSplineProfile(BatchReactor.Tmin,
                                            BatchReactor.Tmin,
                                            0.25)
    end
end
function upper(p :: BatchReactor.TemperatureProfile)
    if typeof(p) == BatchReactor.PiecewiseLinearProfile
        BatchReactor.PiecewiseLinearProfile(ones(length(p.ft)),
                                            ones(length(p.fT)),
                                            BatchReactor.Tmax)
    else
        BatchReactor.QuadraticSplineProfile(BatchReactor.Tmax,
                                            BatchReactor.Tmax,
                                            0.75)
    end
end
domain = Fresa.Domain(lower, upper)
# create initial population consisting of one initial profile for each
# type of representation
p0 = [Fresa.createpoint(BatchReactor.QuadraticSplineProfile(323.0,
                                                            323.0,
                                                            0.5),
                        objective)
      Fresa.createpoint(BatchReactor.PiecewiseLinearProfile([0.5,
                                                             0.5,
                                                             0.5,
                                                             0.5],
                                                            [0.0,
                                                             0.0,
                                                             0.0,
                                                             0.0],
                                                            323.0),
                        objective )]
# invoke Fresa
pareto, population = Fresa.solve(
    # the first 3 arguments are required
    objective,              # the objective function
    p0,                     # an initial point in the design space
    domain;                 # search domain for the decision variables
    # the rest are option arguments for Fresa
    elite = false,          # elitism by default
    fitnesstype = :nondominated,
    ngen = 100,             # number of generations
    npop = 20,               # population size
    nrmax = 5,              # number of runners maximum
    populationoutput = true # output full population every generation
)

println("Pareto (non-dominated) set:")
println("#+plot: ind:1 deps:(2) with:linespoints")
println("$(population[pareto])")
println("Full population:")
println("#+plot: ind:1 deps:(2) with:points")
println("$population")
\end{lstlisting}
\end{document}